\def\year{2021}\relax
\documentclass[letterpaper]{article} % DO NOT CHANGE THIS
\usepackage{aaai20}  % DO NOT CHANGE THIS
\usepackage{times}  % DO NOT CHANGE THIS
\usepackage{helvet} % DO NOT CHANGE THIS
\usepackage{courier}  % DO NOT CHANGE THIS
\usepackage[hyphens]{url}  % DO NOT CHANGE THIS
\usepackage{graphicx} % DO NOT CHANGE THIS
\usepackage{graphicx}  % DO NOT CHANGE THIS
\usepackage[utf8]{inputenc} % allow utf-8 input
\usepackage[T1]{fontenc}    % use 8-bit T1 fonts
\usepackage{booktabs}       % professional-quality tables
\usepackage{amsfonts}       % blackboard math symbols
\usepackage{nicefrac}       % compact symbols for 1/2, etc.
\usepackage{microtype}      % microtypography
\usepackage{xcolor}
\usepackage{subfigure}
\usepackage{xparse}
\usepackage{relsize}
\usepackage{bbm}
\usepackage{stmaryrd}
\usepackage{amsmath}
\usepackage{booktabs} % for professional tables
\usepackage{xspace}
\usepackage[backgroundcolor=white]{todonotes}
\usepackage{multirow}
\usepackage{multicol}
\usepackage{comment}
\usepackage{makecell}

\newcommand{\trail}{TRAIL\xspace}

\usepackage{multirow}
\usepackage{cellspace}
\usepackage{hhline}
\usepackage{stackengine}
\setstackEOL{\\}
\usepackage[switch]{lineno} 
\usepackage{tikz}
\def\checkmark{\tikz\fill[scale=0.4](0,.35) -- (.25,0) -- (1,.7) -- (.25,.15) -- cycle;} 

\DeclareMathOperator*{\argmax}{arg\,max}

\nocopyright

\title{A Deep Reinforcement Learning Approach to First-Order Logic Theorem Proving}
\author{
\normalsize
Maxwell Crouse\thanks{Equal contribution, correspondence to Maxwell Crouse \texttt{<mvcrouse@u.northwestern.edu>}, Ibrahim Abdelaziz \texttt{<ibrahim.abdelaziz1@ibm.com>}, and Achille Fokoue \texttt{<achille@us.ibm.com>}}$^\S$,
Ibrahim Abdelaziz$^{*\ddag}$,
Bassem Makni$^\ddag$,
Spencer Whitehead$^\dag$,
\\
\bf Cristina Cornelio$^\#$, Pavan Kapanipathi$^\ddag$,
 Kavitha Srinivas$^\ddag$,
 Veronika Thost$^\P$, \\
 \bf  Michael Witbrock$^\diamond$,
 Achille Fokoue$^\ddag$
\AND
{\normalsize \rm $^\S$Northwestern University }\\
Department of Computer Science\\
\And
{\normalsize \rm $^\ddagger$IBM Research} \\
IBM T.J. Watson Research Center \\
\And
{\normalsize \rm $^\dagger$University of Illinois at Urbana-Champaign }\\
Department of Computer Science\\
\AND
{\normalsize \rm $^{\P}$MIT-IBM Watson AI Lab}\\
IBM Research
\And
{\normalsize \rm $^{\#}$IBM Research}\\
 IBM Zurich Lab %, R{\"u}schlikon, CH
\And
{\normalsize\rm $^\diamond$The University of Auckland} \\
School of Computer Science \\
}

\begin{document}
%\linenumbers
% For research notes, remove the comment character in the line below.
% \researchnote

\maketitle
% --------------
\begin{abstract}%
Automated theorem provers have traditionally relied on manually tuned heuristics to guide how they perform proof search.
Deep reinforcement learning has been proposed as a way to
obviate the need for such heuristics, however, its deployment
in %state-of-the-art saturation-based
% I took out the saturation-based specification as we present a better approach to deep rl theorem proving in general
automated theorem proving remains a challenge.
In this paper we introduce \trail, a system that
applies deep reinforcement learning to
saturation-based theorem proving. \trail leverages
(a) a novel neural representation of the state of a theorem prover
and (b) a novel characterization of the inference selection process
in terms of an attention-based action policy.
%
%We show through systematic analysis that these mechanisms allow \trail to significantly outperform previous reinforcement-learning-based theorem provers and approach the performance of state-of-the-art traditional theorem provers on two benchmark datasets for first-order logic automated theorem proving. 
We show through systematic analysis that these mechanisms allow \trail to significantly outperform previous reinforcement-learning-based theorem provers on two benchmark datasets for first-order logic automated theorem proving (proving around 15\% more theorems). 
\end{abstract}

\section{Introduction}
Automated theorem provers (ATPs) 
have established themselves as useful tools for solving problems that are expressible in a variety of knowledge representation formalisms. Such problems are commonplace in areas core to computer science (e.g., compilers \cite{curzon1991verified,leroy2009formal}, operating systems \cite{klein2009operating}, and even distributed systems \cite{hawblitzel2015ironfleet,garland1998ioa}), where ATPs are used to prove that a system satisfies some formal design specification.
Unfortunately, while the formalisms that underlie such problems have been (more or less) fixed, the strategies needed to solve them have been anything but. With each new domain added to the purview of automated theorem proving, there has been a need for the development of new heuristics and strategies that restrict or order how an ATP searches for proofs. This process of guiding a theorem prover during proof search is referred to as \emph{proof guidance}. Though proof guidance heuristics have been shown to have a drastic impact on theorem proving performance \cite{schulz2016performance}, the specifics of when and why to use a particular strategy are still often hard to define \cite{schulzwe}.

Many state-of-the-art ATPs use machine learning to automatically determine heuristics for assisting with proof guidance. Generally, the features considered by such learned heuristics have been manually designed \cite{KUV-IJCAI15-features,JU-CICM17-enigma}, though more recently they have been learned through deep learning \cite{LoosISK-LPAR17-atp-nn,ChvaJaSU-CoRR19-enigma-ng,paliwal2019graph}, which has the appeal of lessening the amount of expert knowledge needed compared with handcrafting new heuristics. These neural approaches have just begun to yield impressive results, e.g. Enigma-NG \cite{JU-CoRR19-enigma-hammering} showed that purely neural proof guidance could be integrated into E \cite{schulz2002brainiac} to improve its performance over manually designed proof-search strategies. However, in order to achieve competitive performance with state-of-the-art ATPs, neural methods (as they have been applied thus far) have critically relied on being seeded with proofs from an existing state-of-the-art reasoner (which itself will use a strong manually designed proof-search strategy). Thus, such approaches are still subject to the biases inherent to the theorem-proving strategies used in their initialization.
%However, while deep learning has the clear advantage of being able to lessen the amount of expert knowledge needed to handcraft new heuristics, prior work has focused on bootstrapping their systems with proofs output by existing reasoners. \cite{alphagozero}
%the acquisition of sufficiently large quantities of high quality training data remains a challenge. {\color{red} Highlight differences between deep learning and RL, e.g. deep learning systems bootstrap}

Reinforcement learning \emph{a la} AlphaGo Zero \cite{alphagozero} has been explored as a natural solution to this problem, where the system automatically learns proof guidance strategies from scratch. More generally, reinforcement learning has been successfully applied to theorem proving with first-order logic  \cite{KalUMO-NeurIPS18-atp-rl,piotrowski2019guiding,zombori2019towards,zombori2020prolog}, higher-order logic \cite{BaLoRSWi-CoRR19-learning}, and also with logics less expressive than first-order logic \cite{KuYaSa-CoRR18-intuit-pl,LeRaSe-CoRR18-heuristics-atp-rl,CheTi-CoRR18-rew-rl}. 

Here we target theorem proving for first-order logic, where \emph{tabula rasa} reinforcement learning (i.e., learning from scratch) has been integrated into tableau-based theorem provers \cite{KalUMO-NeurIPS18-atp-rl,zombori2019towards,zombori2020prolog}. 
Connection tableau theorem proving is an appealing setting for machine learning research because tableau calculi are straightforward and simple, allowing for concise implementations 
%(e.g., leanCop's core procedure can be written in seven lines of code \cite{KalUMO-NeurIPS18-atp-rl}) 
that can easily be extended with learning-based techniques. However, the best performing, most widely-used theorem provers to date are saturation theorem provers that implement either the resolution or superposition calculi \cite{vampire2013,schulz2002brainiac}. These provers are capable of much finer-grained management of the proof search space; however, this added power comes at the cost of increased complexity in terms of both the underlying calculi and the theorem provers themselves. 
For neural-based proof guidance to yield any improvements when integrated with highly optimized, hand-tuned saturation-based provers, it must offset the added cost of neural network evaluation with more intelligent proof search. To date, this has not been possible when these neural approaches have been trained from scratch, i.e. when they are \emph{not} bootstrapped with proofs from a state-of-the-art ATP.
%This makes the integration of saturation-based provers with neural-based reasoning an interesting direction for research, as the added computational cost of neural network evaluation (as compared to the efficient but shallow heuristics currently in use) means that a theorem prover would need \emph{intelligently}.
%Both the underlying calculi as well as the theorem provers themselves are very complex,
%(for reference, Vampire has over 73,000 lines of code \cite{hoder2010interpolation}),
%which makes the non-superficial incorporation of deep learning techniques a challenge in both theory and practice.
%Our focus here thus lies in deep reinforcement learning for saturation-based theorem proving.

%In this paper, we introduce \trail (Trial Reasoner for AI that Learns), a theorem proving approach that applies deep reinforcement learning to saturation-based theorem proving to learn proof guidance strategies completely from scratch.
%maps the core ideas of saturation theorem proving into a deep reinforcement learning framework. This allows for a much deeper integration of modern deep learning techniques with the operations of a saturation-based theorem prover than has yet been possible. 
In this paper, we introduce \trail (Trial Reasoner for AI that Learns), a theorem proving approach that applies deep reinforcement learning to saturation-based theorem proving to learn proof guidance strategies completely from scratch.
% I don't say "first" to apply deep-rl to saturation theorem proving because I'm pretty sure there's already a published workshop paper that gets terrible results that technically could claim their work to be concurrently developed
Key to \trail's design is a novel neural representation of the state of a theorem-prover in terms of inferences and clauses, and a novel characterization of the inference selection process in terms of an attention-based action policy. 
The neural representations for clauses and actions that constitute \trail's internal state were based on a careful study of candidate representations, which we describe in this paper.  %{\color{red} describe novel neural representation of state in terms of the distributed representations used for clauses and actions}

% both M2k and MPTP are from Mizar
We demonstrate the performance of \trail on two standard benchmarks  drawn from the Mizar dataset \cite{mizar40for40}: M2k \cite{KalUMO-NeurIPS18-atp-rl} and MPTP2078 \cite{urban2006mptp}, where we show that \trail, when trained from scratch, outperforms all prior reinforcement-learning approaches on these two datasets and approaches the performance of a state-of-the-art ATP on MPTP2078 dataset. 

%To the best of our knowledge, this is the first time that a neural-based theorem-proving system has achieved performance competitive with that of a state-of-the-art theorem prover when trained completely from scratch.

%In addition, we perform experiments that explore alternative training regimens to determine which mechanism provides the best performance: (a) we randomly explore from a tabula rasa state as done in AlphaZero \cite{alphazero} to build a proof guidance system that maximizes exploration, (b) we explore the effectiveness of incorporating proofs from existing reasoners in a specific problem domain, where the system can effectively learn from highly optimized reasoners.
%%%%%%%%%%%%%%%%%%%%%%%%%%%%%%%%%%%%%%%%%%%%%%%%%%%%%%%%%%%%%%%%
%%%%%%%%%%%%%%%%%%%%%%%%%%%%%%%%%%%%%%%%%%%%%%%%%%%%%%%%%%%%%%%%
%%%%%%%%%%%%%%%%%%%%%%%%%%%%%%%%%%%%%%%%%%%%%%%%%%%%%%%%%%%%%%%%

\section{Background}\label{gen_inst}
%%%%%%%%%%%%%%%%%%%%%%%%%%%%%%%%%%%%%%%%%%%%%%%%%%%%%%%%%%%%%%%%
%fol with equality / Conjunctive Normal Form
%%%%%%%%%%%%%%%%%%%%%%%%%%%%%%%%%%%%%%%%%%%%%%%%%%%%%%%%%%%%%%%%

We assume the reader has knowledge of basic first-order logic and automated theorem proving terminology and thus will only briefly describe the terms commonly seen throughout this paper. For readers interested in learning more about logical formalisms and techniques see \cite{thelogicbook,enderton2001mathematical}.

In this work, we focus on first-order logic (FOL) with equality. In the standard FOL problem-solving setting, an ATP is given a \emph{conjecture} (i.e., a formula to be proved true or false), \emph{axioms} (i.e., formulas known to be true), and \emph{inference rules} 
(i.e., rules that, based on given true formulas, allow for the derivation of new true formulas).
From these inputs, the ATP performs a \emph{proof search}, which can be characterized as the successive application of inference rules to axioms and derived formulas until a sequence of derived formulas is found that represents a \emph{proof} of the given conjecture. All formulas considered by \trail are in \emph{conjunctive normal form}. That is, they are conjunctions of {\it clauses}, which are themselves disjunctions of literals. Literals are (possibly negated) formulas that otherwise have no inner logical connectives. In addition, all variables are implicitly universally quantified.

Let $F$ be a set of formulas and $\mathcal{I}$ be a set of inference rules. We write that $F$ is \emph{saturated} with respect to $\mathcal{I}$ if every inference that can be made using axioms from $\mathcal{I}$ and premises from $F$ is also a member of $F$, i.e. $F$ is closed under inferences from $\mathcal{I}$. Saturation-based theorem provers aim to saturate a set of formulas with respect to their inference rules. To do this, they maintain two sets of clauses, referred to as the \emph{processed} and \emph{unprocessed} sets of clauses. These two sets correspond to the clauses that have and have not been yet selected for inference. The actions that saturation-based theorem provers take are referred to as \emph{inferences}. Inferences involve an inference rule (e.g. resolution, factoring) and a non-empty set of clauses, considered to be the \emph{premises} of the rule. At each step in proof search, the ATP selects an inference with premises in the unprocessed set (some premises may be part of the processed set) and executes it. This generates a new set of clauses, each of which is added to the unprocessed set. The clauses in the premises that are members of the unprocessed set are then added to the processed set. This iteration continues until a clause is generated (typically the empty clause for refutation theorem proving) that signals a proof has been found, the set of clauses is saturated, or a timeout is reached.
% It is not clear, given this exposition, why any inference rule should generate an empty clause, and why the generation of that empty clause would constitute a signal that a proof has been found. More simply, deriving the conjecture would constitute a signal that a proof has been found. 
For more details on saturation \cite{robinson1965machine} and saturation-calculi, we refer the reader to \cite{bachmair1998equational}.

\begin{figure*}[t]
\begin{center}
% \centerline{\includegraphics[width=\columnwidth]{figures/fig1.pdf}}
\includegraphics[width=0.7\textwidth]{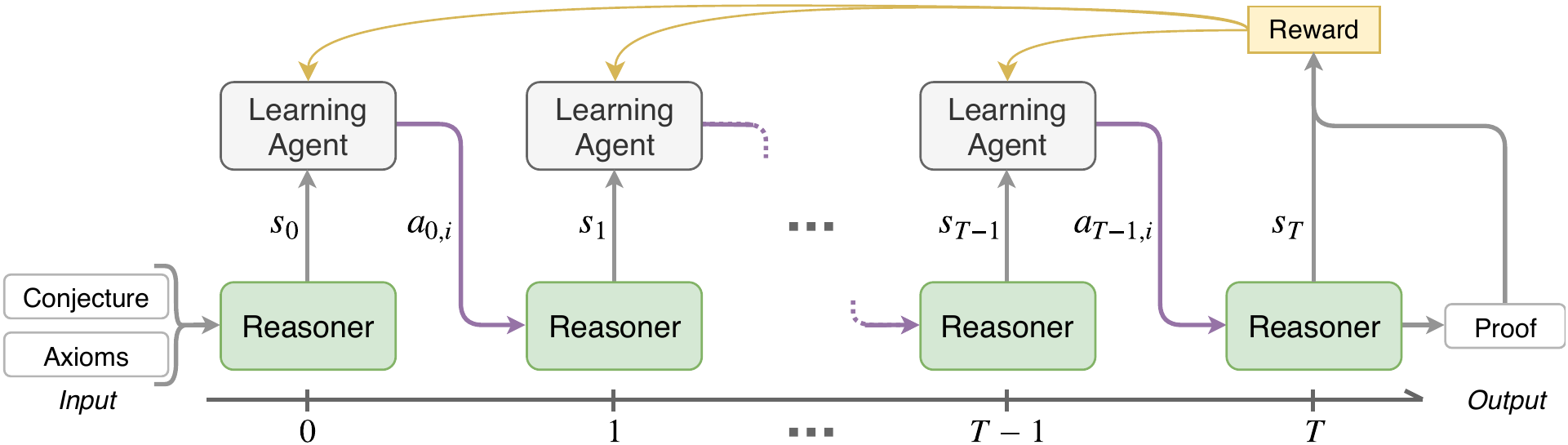}
\caption{Formulation of automated theorem proving as a RL problem}
\label{fig:rl_formulation}
\end{center}
\end{figure*}

%Saturation (introduced in \cite{robinson1965machine}) is only a proof-search procedure, and a straightforward implementation of it within a theorem prover holds no guarantees of efficiency. Much of the power of saturation-based theorem provers instead comes from saturation calculi (a light overview can be found in \cite{bachmair1998equational}); with the most commonly used being the resolution \cite{robinson1965machine} and superposition \cite{bachmair1994rewrite} calculi. These calculi (and their many variants) avoid much of the proof search space explosion by leveraging restrictions on inference-rule applicability with selection functions and simplification orderings \cite{bachmair2001resolution} that maintain completeness. In addition, they apply significant effort to recognizing and eliminating redundancy (e.g., when a generated clause is a logical consequence of some other simpler clause). This can involve discarding the result of an inference (i.e., not adding a consequence to the unprocessed set) or the deletion of some preexisting clause (e.g., demodulation is an inference rule that deletes one of its premises). Redundancy handling can make the state of a theorem prover very dynamic, as the set of clauses under consideration can expand or contract depending on the inferences performed.

\section{{\trail}}\label{sec:approach}
% In this section, we present our approach \trail. Our theorem prover diverges from others in how proof-search is executed, in particular we formulate it as an RL problem. In what follows we provide a detailed description of \trail and specify our training procedure.
%%%%%%%%%%%%%%%%%%%%%%%%%%%%%%%%%%%%%%%%%%%%%%%%%%%%%%%%%%%%%%
We first describe our overall approach to defining the proof guidance problem in terms of reinforcement learning. For this, we detail (a) a \textit{sparse vectorization process} which represents all clauses and actions in a way compatible with subsequent neural components, (b) a \textit{neural proof state} which concisely captures the neural representations of clauses and actions within a proof state, and (c) an \textit{attention-based policy network} that learns the interactions between clauses and actions to select the next action.  Last, we describe how \trail learns from scratch, beginning with a random initial policy.
\subsection{Reinforcement Learning for Proof Guidance}
% \subsection{Problem Formulation}
\label{sec:rl_formulation}

%As illustrated in Figure~\ref{fig:rl_formulation}, 
% We formalize automated theorem-proving (ATP) as an RL problem where the reasoner provides the environment in which the learning agent operates.
We formalize the proof guidance as a reinforcement learning (RL) problem where the reasoner provides the environment in which the learning agent operates.
Figure~\ref{fig:rl_formulation} shows how an ATP problem is solved in our framework.
%where environment is a classical reasoner whose actions are dictated by the learning agent. 
%If there is space, the caption of Figure 1 should be improved to mimimally explain the process that is illustrated in the figure, so that it can be understood in isolation. Its meaning is currently opaque without looking at the running text.
Given a conjecture and a set of axioms, \trail iteratively performs reasoning steps until a proof is found (within a provided time limit).
The reasoner tracks the proof state, $s_t$, which encapsulates the clauses that have been derived or used in the derivation so far and the actions that can be taken by the reasoner at the current step.
At each step, this state is passed to the learning agent - an attention-based model~\cite{luong2015attention} that predicts a distribution over the actions it uses to sample a corresponding action, $a_{t,i}$. This action is given to the reasoner, which executes it and updates the proof state.
% (i.e., a network \dots \cite{}; see also Section~\ref{sec:policy-network})

Formally, a state, $s_{t} = (\mathcal{C}_{t}, \mathcal{A}_{t})$, consists of two components. The first is the set of processed clauses, $\mathcal{C}_{t} = \{c_{t,j}\}_{j=1}^{N}$,  (i.e., all clauses selected by the agent up to step $t$); where $\mathcal{C}_{0} = \emptyset$. The second is the set of all available actions that the reasoner could execute at step $t$, $\mathcal{A}_{t} = \{a_{t,i}\}_{i=1}^{M}$; where $\mathcal{A}_{0}$ is the cross product of the set of all inference rules (denoted by $\mathcal{I}$) and the set of all axioms and the negated conjecture. An action, $a_{t,i} = (\xi_{t,i}, \hat{c}_{t,i})$, is a pair comprising an inference rule, $\xi_{t,i}$, and a clause from the unprocessed set, $\hat{c}_{t,i}$.% This $\hat{c}_{t,i}$ is either an axiom, the negated conjecture, or a derived clause yet to be selected by the agent. %Informally, $\mathcal{A}_{t}$ represents the unprocessed clauses at step $t$ and the inference rules applicable to them. 

At step $t$, given a state $s_{t}$ (provided by the reasoner), the learning agent computes a probability distribution over the set of available actions $\mathcal{A}_t$, denoted by $P_\theta(a_{t,i} | s_{t})$ (where $\theta$ is the set of parameters for the learning agent), and samples an action $a_{t,i} \in \mathcal{A}_t$. The sampled action $a_{t,i} = (\xi_{t,i}, \hat{c}_{t,i})$ is executed by the reasoner by applying $\xi_{t,i}$ to $\hat{c}_{t,i}$ (which may involve processed clauses). This yields a set of new derived clauses, $\bar{\mathcal{C}}_{t}$, and a new state, $s_{t+1} = (\mathcal{C}_{t+1}, \mathcal{A}_{t+1})$, where $\mathcal{C}_{t+1} = \mathcal{C}_{t}\cup \{\hat{c}_{t,i}\}$ and $\mathcal{A}_{t+1} = (\mathcal{A}_{t} - \{a_{t,i}\}) \cup (\mathcal{I} \times \bar{\mathcal{C}}_{t})$.

Upon completion of a proof attempt, \trail computes a loss and issue a reward that encourages the agent to optimize for decisions leading to a successful proof in the shortest time possible. 
% in the smallest number of steps. 
Specifically, for an unsuccessful  proof attempt (i.e., the underlying reasoner fails to derive a contradiction within the time limit), each step $t$ in the attempt is assigned a reward $r_t=0$. For a successful proof attempt, in the final step, the underlying reasoner produces a refutation proof $\mathcal{P}$ containing only the actions that generated derived facts directly or indirectly involved in the final contradiction. At step $t$ of a successful proof attempt where the action $a_{t, i}$ is selected, the reward $r_t$ is $0$ if $a_{t, i}$ is not part of the refutation proof $\mathcal{P}$; otherwise $r_t$ is inversely proportional to the time spent proving the conjecture. %the total number of steps in the proof attempt.

The final loss consists of the standard policy gradient loss~\cite{sutton1998reinforcement} and an entropy regularization term to avoid collapse onto a sub-optimal deterministic policy and to promote exploration.  
\begin{alignat*}{2}\label{eqn:loss}
       &\mathcal{L}(\theta) = &&-\mathbb{E}\big[r_t \log(P_{\theta}(a_{t} | s_{t})) \big] \\
       & &&-\lambda \mathbb{E}\big[\sum_{i=1}^{|\mathcal{A}_t|} - P_{\theta}(a_{t,i} | s_{t}) \log(P_{\theta}(a_{t,i} | s_{t})) \big]
\end{alignat*}
where $a_t$ is the action taken at step $t$ and $\lambda$ is the entropy regularization hyper-parameter. We use a normalized reward to improve training stability, since the intrinsic difficulty of problems can vary widely in our problem dataset. We explored (i) normalization by the inverse of the time spent by a traditional reasoner, (ii) normalization by the best reward obtained in repeated attempts to solve the same problem, and (iii) no normalization; the normalization strategy was a hyper-parameter.
This loss has the effect of giving actions that contributed to the most direct proofs for a given problem higher rewards, while dampening actions that contributed to more time consuming proofs for the same problem.

%%%%%%%%%%%%%%%%%%%%%%%%%%%%%%%%%%%%%%%%%%%%%%%%%%%%%%%%%%%%%%
%\subsection{Generalized Learning Agent}
%The vectorizer computes vector representations of clauses and actions in a given proof state in a way that abstracts away the specifics of the vocabulary of a problem. The policy network uses these representations along with a neural attention mechanism to learn the meaningful interactions between clauses and actions to select the action for the underlying reasoner to execute.

\subsection{Sparse Vectorization Process}
\label{sec:vectorization}
%\subsubsection{Producing Sparse Representations}
%For compatibility with subsequent neural components, all clauses and inferences must first be transformed into real-valued, fixed-length vectors.
For compatibility with subsequent neural components, \trail transforms the first-order logic forms internal to the proof state into real-valued vectors.
To do this, \trail utilizes a set of $M$ vectorization modules, $\mathcal{M} = \{ m_1, \ldots, m_M \}$, that each characterize some important aspect of the clauses and actions under consideration.

Each module $m_k \in \mathcal{M}$ follows the same general design: given an input clause or action, $m_k$ produces a discrete, bag-of-words style vector in $\mathbb{Z}^{n_k}$, where $n_{k}$ is a pre-specified dimensionality specific to module $m_k$. As an example, consider a module intended to capture the symbols of a formula. It would map each symbol present in its input to an index ranging from $1, \ldots, n_k$. The vector representation would be created by assigning to each dimension of a sparse $n_k$-dimensional vector the number of times a symbol corresponding to that dimension appeared in the input (i.e., a bag-of-words representation). Letting $m_k(i)$ be the sparse vector for an input $i$ from module $m_k \in \mathcal{M}$, the final vector representation $v_i$ is then the concatenation of the outputs from each module. 
% , i.e. $$ v_i = \bigparallel_{k = 1}^{M} m_k(i) ~.$$
% I removed the formula above since is obvious what we mean with "the concatenation of the outputs"
As actions are pairs of clauses and inference rules, \trail represents the clause in each action pair using the process described above and the inference rule as a one-hot encoding of size $|\mathcal{I}|$, where $\mathcal{I}$ is the set of inference rules.

This vectorization process allows \trail to trivially incorporate techniques from the large body of research on vectorization strategies for machine learning guided theorem proving \cite{bridge2014machine,KUV-IJCAI15-features}. For instance, \trail uses Enigma \cite{JU-CICM17-enigma} modules which characterize a clause in terms of fixed-length term walks (with separate modules for term walks of length $l \in \{ 1, 2, 3\}$). In addition to term walks, \trail uses general purpose modules for representing clause age, weight, literal count, and set-of-support features, as well as one new feature type described below.

%{\color{red} describe GCN formulation}

\begin{figure}[b]
\begin{center}
\includegraphics[width=1.0\columnwidth]{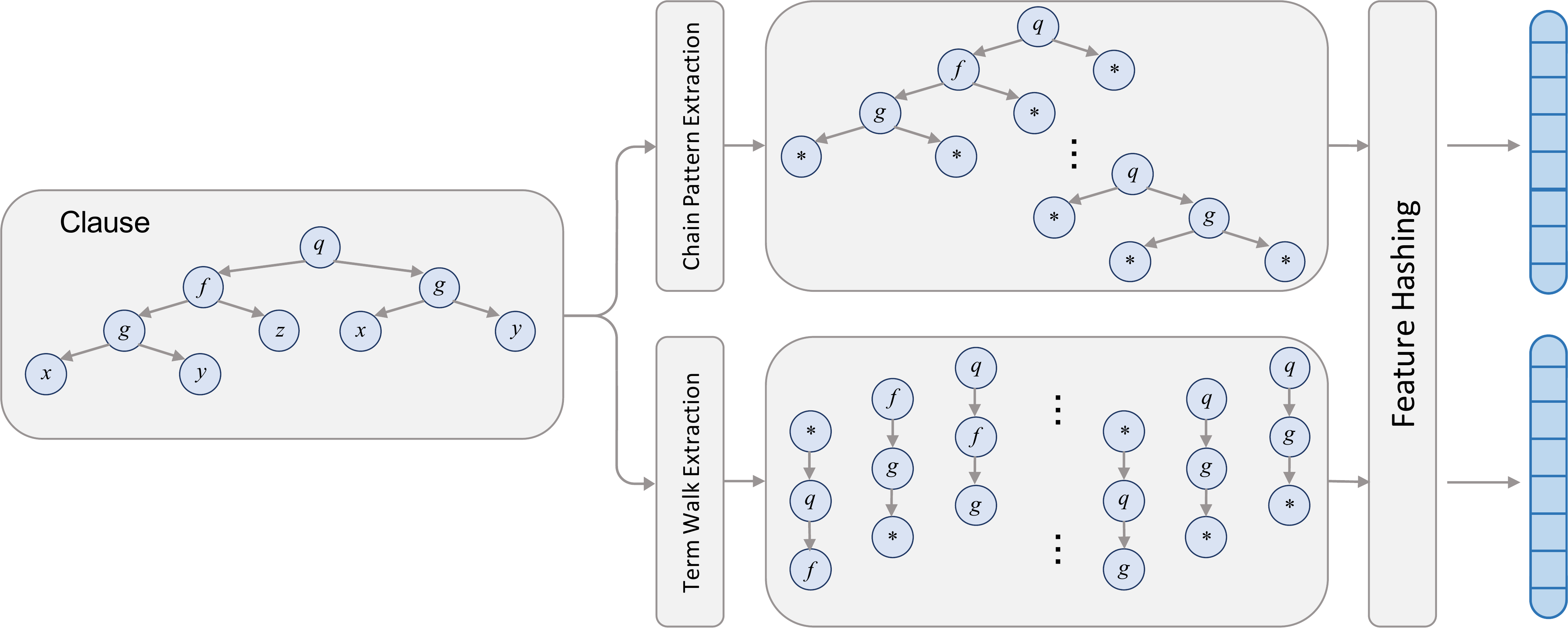}
\caption{Overview of chain and term walk vectorizers operating on the clause $q(f(g(x, y), z), g(x, y))$}
\label{fig:herbrand_templates}
\end{center}
\end{figure}

\subsubsection{Chain-Based Vectorization}
%Figure~\ref{fig:herbrand_templates} shows an example of the vectorizer operating on a clause with a single positive literal. 
We refer to the vectorization method introduced here as \emph{chain-based vectorization}. Within \trail, it functions as one of the available vectorization modules. Its development was inspired by the way inference rules operate over clauses. Consider the resolution inference rule. Two clauses resolve if one contains a positive literal whose constituent atom is unifiable with the constituent atom of a negative literal in the other clause.
Hence, vector representations of clauses should both capture the relationship between literals and their negations and reflect structural similarities between literals that are indicative of unifiability.

Our approach captures these features by deconstructing input clauses into sets of patterns. We define a \textit{pattern} to be a linear chain that begins from a predicate symbol and includes one argument (and its argument position) at each depth until it ends at a constant or variable. The set of all patterns for a clause is then the set of all linear paths between each predicate and the constants and variables they bottom out with. Since the names of variables are arbitrary, they are replaced with a wild-card symbol, ``$\ast$''.  %indicating that the element may match with anything. 
Argument position is also indicated with the use of wild-card symbols.
%Going back to the clause in Figure~\ref{fig:herbrand_templates}, 
As an example, for the single literal clause $q(f(g(x, y), z), g(x, y))$, we would obtain the patterns $q(f(g(\ast, \ast), \ast), \ast)$ and $q(\ast, g(\ast, \ast))$). 

We obtain a $d$-dimensional representation of a clause by hashing the linearization of each pattern $p$ using MD5 hashes~\cite{rivest1992md5} to compute a hash value $v$, and setting the element at index $v \bmod d$ to the number of occurrences of the pattern $p$ in the clause. We also explicitly encode the difference between patterns and their negations by doubling the representation size and hashing them separately, where the first $d$ elements encode the positive patterns and the second $d$ elements encode the negated patterns. Chain vectorization is intended to produce feature vectors useful for estimates of structural similarity. Figure \ref{fig:herbrand_templates} shows how chain patterns would be extracted from a clause in along with term walks.

An important design decision for \trail is to rely on a combination of simple and efficient vectorizers like the pattern-based vectorizers described above, as opposed to more complex graph neural network (GNN) approaches. Though GNN-based  approaches are powerful, they introduce a noticeable additional run-time overhead (i.e., more time spent looking for a better search strategy and less time left to execute it). We contrast our approach with a graph neural network based approach in Section~\ref{sec:experiments}.

\begin{figure*}[t]
\begin{center}
\includegraphics[width=0.71\textwidth]{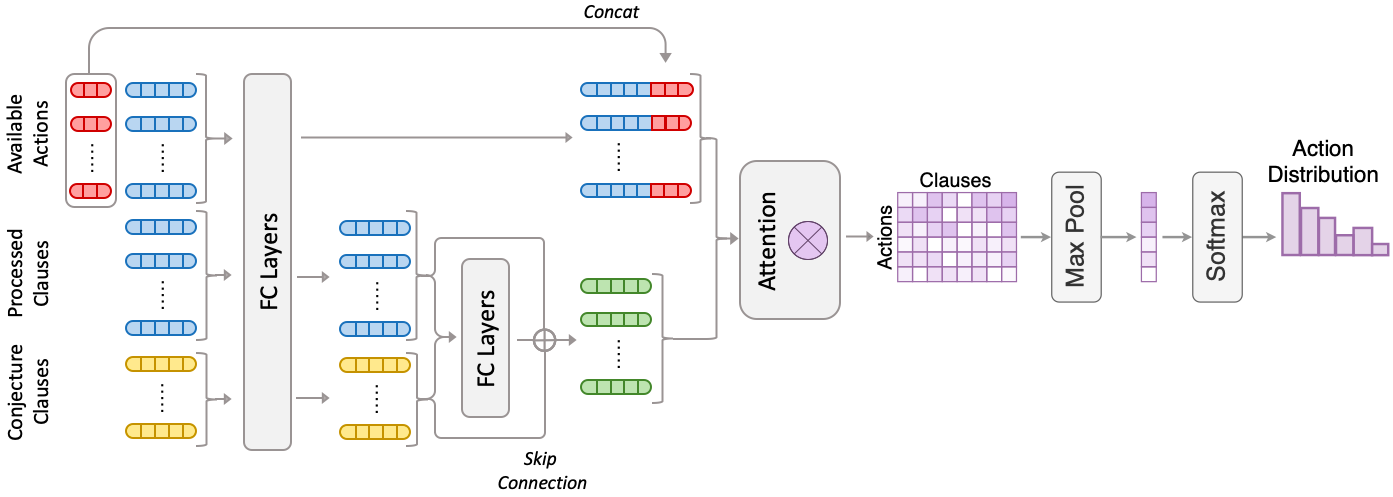}%figures/trail_policy_net_final.pdf}
\caption{Flow from sparse vectorization through the policy network}
\label{fig:network_arch}
\end{center}
\end{figure*}

\subsection{Neural Representation of Proof State}

Recall that the proof state consists of the sets of processed clauses $\mathcal{C}_t$ and actions $\mathcal{A}_t$. We write $\mathbf{c}_{t,i}$ as the sparse vector representation for processed clause $c_i$ at time $t$ and $(\mathbf{z}_{t,i}, \hat{\mathbf{c}}_{t,i})$ as the vector representations for the inference rule and clause pairing for action $a_{t,i}$. 
%The set of processed clause representations can then be written as $\{\mathbf{c}_{t,1},...,\mathbf{c}_{t,N}\}$ and the set of action representations as $\{(\mathbf{z}_{t,1}, \hat{\mathbf{c}}_{t,1}),..., (\mathbf{z}_{t,M}, \hat{\mathbf{c}}_{t,M})\}$.
To produce dense representations for the elements of $\mathcal{C}_t$ and $\mathcal{A}_t$, \trail first transforms the sparse clause representations (excluding inference rules for actions) %(from the set of processed clauses or actions) 
into dense representations by passing them through $k$ fully-connected layers. This yields sets $\{\mathbf{h}^{(p)}_{t,1},..., \mathbf{h}^{(p)}_{t,N}\}$ and $\{\mathbf{h}^{(a)}_{t,1},..., \mathbf{h}^{(a)}_{t,M}\}$ of dense representations for the processed and action clauses. \trail also collects the dense representations for the negated conjecture clauses as $\{\mathbf{h}^{(c)}_{1}, \ldots \mathbf{h}^{(c)}_k\}$.

For each action pair, \trail concatenates the clause representation $\mathbf{h}^{(a)}_{t,i}$ with the corresponding inference representation $\mathbf{z}_{t,i}$ to form the new action $\mathbf{a}_{t,i} = [\mathbf{h}^{(a)}_{t,i}, \mathbf{z}_{t,i}]$ and joins each such action into the matrix $\mathbf{A}$.
To construct the processed clause matrix, \trail first produces a dense representation of the conjecture as the element-wise mean of dense negated conjecture clause representations
\begin{align*}
\mathbf{h}^{(c)} = \dfrac{1}{k}\sum_{i = 1}^k \mathbf{h}^{(c)}_i
\end{align*}
where $k$ is the number of conjecture clauses. New processed clause representations are produced by combining the original dense representations with the pooled conjecture. For a processed clause embedding $\mathbf{h}^{(p)}_i$, its new value would be
\begin{align*}
\hat{\mathbf{h}}^{(p)}_i = \mathbf{h}^{(p)}_i + \mathbf{h}^{(c)} + F(\mathbf{h}^{(p)}_i \ || \  \mathbf{h}^{(c)})
\end{align*}
where \textit{F} is a feed-forward network, $||$ denotes the concatenation operation, and the original inputs are included with skip connections \cite{he2016deep}. The new processed clause embeddings are then joined into the matrix $\mathbf{C}$.

The two resulting matrices $\mathbf{C}$ and $\mathbf{A}$ can be considered the neural forms of $\mathcal{C}_t$ and $\mathcal{A}_t$. Thus, they concisely capture the notion of a \emph{neural proof state}, where each column of either matrix corresponds to an element from the formal proof state. Following the construction of $\mathbf{C}$ and $\mathbf{A}$, this neural proof state is fed into the policy network to select the next inference.

\subsection{Attention-Based Policy Network}\label{sec:policy-network}

% \textbf{Policy Network.}
Figure~\ref{fig:network_arch} shows how sparse representations are transformed into the neural proof state and passed to the policy network. Throughout the reasoning process, the policy network must produce a distribution over the actions relative to the clauses that have been selected up to the current step, where both the actions and clauses are sets of variable length.
This setting is analogous to ones seen in attention-based approaches to problems like machine translation~\cite{luong2015attention,vaswani2017attention} and video captioning~\cite{yu2016video,whitehead2018kavd}, in which the model must score each encoder state with respect to a decoder state or other encoder states.
To score each action relative to each clause, we compute a multiplicative attention~\cite{luong2015attention} as
$\mathbf{H} = \mathbf{A}^\top \mathbf{W}_a \mathbf{C}$, 
% \begin{equation*}
% \mathbf{H} = \mathbf{A}^\top \mathbf{W}_a \mathbf{C} ~, 
% \end{equation*}
where $\mathbf{W}_a \in \mathbb{R}^{(2d + |\mathcal{I}|) \times 2d}$ is a learned parameter and the resulting matrix, $\mathbf{H} \in \mathbb{R}^{M \times N}$, is a heat map of interaction scores between processed clauses and available actions. \trail then performs max pooling over the columns (i.e., clauses) of $\mathbf{H}$ to find a single score for each action and apply a softmax normalization to the pooled scores to obtain the distribution over the actions, $P_\theta(a_{t,i} | s_{t})$.

Prior work integrating deep learning with saturation-based ATPs would use a neural network to score the unprocessed clauses with respect to \emph{only} the conjecture and \emph{not} the processed clauses \cite{LoosISK-LPAR17-atp-nn,JU-CoRR19-enigma-hammering}. \trail's attention mechanism can be viewed as a natural generalization of this, where inference selection takes into account both the processed clauses and conjecture.

%%%%%%%%%%%%%%%%%%%%%%%%%%%%%%%%%%%%%%%%%%%%%%%%%%%%%%%%%%%%%%%

%%%%%%%%%%%%%%%%%%%%%%%%%%%%%%%%%%%%%%%%%%%%%%%%%%%%%%%%%%%%%%%
%\subsection{Training Regimens}\label{sec:train}
\subsection{Learning From Scratch}\label{sec:train}

%We test two different training strategies to determine which yields the best overall performance.
%\subsubsection{Random Exploration}
\trail begins learning through random exploration of the search space as done in AlphaZero \cite{alphazero} to establish performance when the system is started from a \textit{tabula rasa} state (i.e., a randomly initialized policy network $P_\theta$).  At training, at an early step $t$ (i.e., $t < \tau_0$, where $\tau_0$, the temperature threshold, is a hyper-parameter that indicates the depth in the reasoning process at which training exploration stops), we sample the action $a_{t,i}$ in the set of available actions  $\mathcal{A}_{t}$, according to the following probability distribution $\widehat{P}_\theta$ derived from the policy network $P_\theta$:
\begin{align*}
    \widehat{P}_\theta(a_{t,i} | s_{t}) = \frac{P_\theta(a_{t,i} | s_{t})^{1/\tau}}{\mathlarger\sum_{a_{t,j} \in \mathcal{A}_{t} }{P_\theta(a_{t,j} | s_{t})^{1/\tau}} }
\end{align*}
where $\tau$, the temperature, is a hyperparameter that controls the exploration-exploitation trade-off and decays over the iterations (a higher temperature promotes more exploration). When the number of steps already performed is above the temperature threshold (i.e., $t \geq \tau_0$), an action, $a_{t,i}$, with the highest probability from the policy network, is selected, i.e.
\begin{align*}
a_{t,i} = \argmax_{a_{t,j} \in \mathcal{A}_{t} } P_\theta(a_{t,j} | s_{t}) ~. 
\end{align*}    
At the end of training iteration $k$, the newly collected examples and those collected in the previous $w$ iterations ($w$ is the example buffer hyperparameter) are used to train, in a supervised manner, the policy network using the reward structure and loss function defined in Section~\ref{sec:rl_formulation}.

\section{Experiments and Results}
\label{sec:experiments}

In this section, we are trying to answer the following questions:
(a) Is \trail effective at generalized proof guidance? For this, we tested whether \trail's performance was competitive with prior ATP systems, both traditional and learning-based.
(b) What is the best vectorization strategy and how well does it generalize / transfer to unseen problems?
\subsubsection{Datasets}
We evaluated {\trail} on two datasets: \emph{M2k}~\cite{KalUMO-NeurIPS18-atp-rl} and \emph{MPTP2078}~\cite{alama2014premise}.  Both datasets are subsets of Mizar\footnote{\url{https://github.com/JUrban/deepmath/}} dataset~\cite{mizar}  which is a well-known mathematical library of formalized 
%and mechanically verified % COMMENT: not all of them
mathematical problems. The M2k dataset contains 2003 problems selected randomly from the subset of Mizar that is known to be provable by existing ATPs, while MPTP2078 
%is another set of 
contains 2078 problems %, however problems in this set were 
selected regardless of whether or not they could be solved by an ATP system.

\subsection{Effectiveness of \trail}

For traditional ATP systems, we compare \trail to: 1) E-prover~\cite{schulz2002brainiac} in auto mode, a state-of-the-art saturation-based ATP system that has been under development for over two decades, 2) Beagle~\cite{Beagle2015}, a newer saturation-based theorem prover that has achieved promising (but not yet state-of-the-art) results in recent ATP competitions, and 3) mlCop \cite{kaliszyk2015certified}, an OCaml reimplementation of leanCop \cite{otten2003leancop}, which is a connection-based refutational theorem prover that was applied to M2k in \cite{KalUMO-NeurIPS18-atp-rl} and MPTP2078 in \cite{zombori2020prolog}. For learning-based approaches, we compare against two recent reinforcement-learning-based approaches: rlCop ~\cite{KalUMO-NeurIPS18-atp-rl} and plCop~\cite{zombori2020prolog}, both of which are connection tableau-based theorem provers that build off mlCop and leanCop, respectively. 
All results for mlCop, rlCop, plCop, and E are those reported in \cite{KalUMO-NeurIPS18-atp-rl,zombori2020prolog}, except for E on M2k which we ran ourselves. We also replicated plCop and rlCop numbers under our exact hardware and time constraints and found comparable results.   
We report all the hyperparameters for \trail in Appendix A.1
along with an overview of the underlying reasoner (Beagle) \trail uses to execute its selected actions (Appendix A.2).
%We also describe in Appendix \ref{sec:underlying_reasoner} the underlying reasoner (Beagle) \trail uses to execute its selected actions.

Following \cite{KalUMO-NeurIPS18-atp-rl,zombori2020prolog}, we report the \textit{best iteration completion performance}. In this setting, \trail starts from scratch and is  applied to M2k and MPTP2078 for 10 iterations, with learning from solved problems occurring after each iteration completes. The reported number of problems solved is then the number solved in the best performing iteration.

\begin{table}[t]
 \centering
\footnotesize
\begin{tabular}{l l c c}
\toprule
 & & M2k & MPTP2078  \\
\toprule
% \multirow{3}{*} {\setstackgap{S}{4.05ex}\Centerstack[l]{Traditional}}&Beagle~\cite{Beagle2015} & 1,543 & 742 \\
% & E~\cite{schulz2002brainiac} & 1,922 &  998\\
\multirow{3}{*} {\setstackgap{S}{4.05ex}\Centerstack[l]{Traditional}}& E & \bf 1922 &  \bf 998\\
&Beagle & 1543 & 742 \\
& mlCop & 1034 & 502\\

\midrule
\multirow{3}{*} {\setstackgap{S}{4.05ex}\Centerstack[l]{Reinforcement \\ Learning-Based}}& rlCop &  1235 & 733 \\
& plCop & 1359 & 782 \\
% &\trail & 1,520 & 786 \\
% &\trail (w/ conjecture) & 1,459 & 819 \\
&\trail  & \bf{1561} & \bf{910} \\
\bottomrule
\end{tabular}
\caption{Number of problems solved in M2k and MPTP2078. %TRAIL numbers are with 100 seconds time limit. 
Best two approaches in \textbf{bold}
}
\label{tab:m2k_2078}
\end{table}

Table~\ref{tab:m2k_2078} shows the performance of \trail against both traditional ATP systems and recent learning-based approaches. Following~\cite{zombori2020prolog}, we limit TRAIL to a maximum of 2,000 steps per problem with a hard limit of 100 seconds. As Table~\ref{tab:m2k_2078} shows, while \trail outperformed Beagle, state-of-the-art saturation-based ATP systems are still superior to RL-based systems, with E solving the most problems on both datasets. Among learning-based approaches, \trail\ solved 202 more problems on M2k compared to plCop~\cite{zombori2020prolog}, and on MPTP2078, \trail\ solved 128 more problems. On both datasets \trail outperformed its underlying reasoner, Beagle, with a 22\% relative improvement on MPTP2078. These results show that \trail (when trained from a random initial policy) is a competitive theorem proving approach, as it outperformed all other systems except for E, with significant gains over prior RL-based approaches. Notably, \trail substantially narrows the gap between state-of-the-art traditional and from-scratch RL approaches on MPTP2078, which includes problems from the non-ATP-solvable portion of Mizar.

% set iterations to 20
%------------on both M2078b and M2k datasets-----

\subsubsection{Value of Attention Over Processed Clauses}

As the conjecture is the end goal of proof search, it is intuitive that it is of key importance to inference selection. However, we would expect processed clauses to provide useful information as well, since most inferences will involve premises from both the unprocessed and processed clauses. To determine the utility of including processed clauses in action selection, we performed an ablation experiment that restricted \trail to utilizing \emph{only} the conjecture's neural representation in the processed clause matrix (i.e., the embedded representation for each processed clause became $\hat{\mathbf{h}}^{(p)}_i = \mathbf{h}^{(c)}$). This turned the attention mechanism into a measure of how aligned an action was to solely the conjecture representation.

When using only the conjecture representations, \trail solved 1,561 problems on M2k and 854 problems on MPTP. Recall that \trail when incorporating processed clause and conjecture representations solved 1,561 problems on M2k and 910 problems on MPTP. Though there was no difference for M2k, it is significant that \trail could solve 56 more problems ($\sim$6\% relative improvement) on MPTP when incorporating processed clauses. 
%It shows that, in certain cases, performance is being left on the table 

% \usepackage{multirow}
\begin{table*}[t]
\centering
\footnotesize
\begin{tabular}{llcc cc}
\toprule
                                  &                      & \multicolumn{2}{c}{No Transfer} & \multicolumn{2}{c}{Transfer}                  \\
\cmidrule(l{2pt}r{2pt}){3-4}\cmidrule(l{2pt}r{2pt}){5-6}
                                  &                      & M2k               & MPTP   & MPTP $\rightarrow$ M2k & M2k $\rightarrow$ MPTP  \\
% \cmidrule{3-6}
\midrule
% \multirow{3}{*}{\begin{tabular}[c]{@{}l@{}}~~~~~Sparse \\ Vectorization\end{tabular}}           & Chains               &   1473	     &   653       &    1463      &    418  \\
\multirow{3}{*}{Sparse Only}           & Chains               &   1473	     &   653       &    1463      &    418  \\

                                  & Term Walks                &  1552	     &   737        &   1425      &     570 \\
                                  & Chains + Term Walks       &  \bf 1561         &  \bf 910  &  \bf 1510     &   \bf 812\\
\midrule
Neural  Only                   & GCN                  &  1523	     &   750       &   1394     &     712 \\
\midrule
\multirow{3}{*}{Sparse \& Neural} &  GCN + Chains         &    1495          &    734      &    1395  &   652     \\
                                  & GCN + Term Walks          &    1549	    &     660     &   1394       &    611         \\
                                  &  GCN + Chains + Term Walks &    1555	    &    706      &   1435       &  690       \\
\bottomrule
\end{tabular}
\caption{TRAIL's performance with different vectorization methods in terms of number of proved theorems. Best results in \textbf{bold}}
\label{tab:vect_transfer}
\end{table*}

\subsection{Effects of Vectorization Choice}

The term walk and chain-based vectorization modules used by \trail  are intended to quickly extract structural features from their inputs. Though they are not tailored to any particular domain within FOL, they are clearly feature engineering. Graph neural network (GNN) approaches have been proposed as a general alternative that can lessen the need for expert knowledge by having a neural network automatically learn salient features from the underlying graph representations of logical formulas. Though GNN approaches have quickly gained traction in this domain due to the graph-centric nature of automated reasoning \cite{paliwal2019graph,WangTWD-NIPS17-deepgraph,olvsak2019property}, they have not yet clearly demonstrated that the value they provide in terms of formula representation offsets their greater computational cost when deployed in a comparison with a state-of-the-art FOL ATP.

We view \trail as providing a framework for further testing this, as GNN-based vectorizers can be trivially substituted into \trail's operation as one of its initial vectorization modules (see Section \ref{sec:vectorization}) that feed into the computation of the neural proof state. Consequently, 
%To determine the effectiveness of the chain-based vectorization method introduced in Section \ref{sec:vectorization},
we performed an experiment where we varied the vectorization modules available to \trail, comparing four different ablated versions: 1) \trail using the chain-based vectorizer, 2) \trail using the term walk vectorizer \cite{JU-CICM17-enigma}, 3) \trail using a GCN-based vectorizer~\cite{kipf2016semi}, and 4) combinations of these vectorizers. For all ablations, \trail had access to the general-purpose vocabulary-independent features (e.g., age, literal count, etc.) mentioned in Section \ref{sec:vectorization}. The GCN is the standard GCN described in \cite{kipf2016semi} operating over the shared subexpression forms of clauses \cite{WangTWD-NIPS17-deepgraph}.

We report \trail's performance using both the prior \emph{best iteration performance} metric (where the train and test sets are the same) and a more standard evaluation where the train and test sets are distinct. For the latter evaluation, we use one dataset (i.e., M2k or MPTP) to train and the other to test (see next Section). Table~\ref{tab:vect_transfer} shows the performance of \trail for these two metrics on both datasets, with the distinction between the sparse vectorization modules, the GNN-based module, and all combinations of each explicitly indicated.

\subsubsection{Performance Within Datasets}
Examining Table \ref{tab:vect_transfer}, we see that while chain-based vectorization is weaker than term walk vectorization, the combination of the two achieves significantly better results as a whole. This suggests that they are capturing non-redundant feature sets which is ideal. Additionally, both the term-walk vectorization and the GCN are very strong on M2k, but they do not do well in every case (i.e., there seems to be some important feature they not capturing). Notably, while the GCN alone provides quite reasonable performance, combining it with other vectorization modules leads to a decrease in performance; which we suspect is due to the more expensive overall computational cost. However, that the GCN provides a strong baseline is promising, as it required no domain-expert knowledge (unlike the term walk features). Our overall takeaway from this experiment is that cheap vectorization methods are currently still quite useful; however, the gap between those techniques and GNN-based approaches is closing.

\subsubsection{Generalization Across Datasets}
\label{sec:transfer}
Table~\ref{tab:vect_transfer} also shows the performance of \trail when trained on M2k and tested on MPTP dataset, and vice versa. In terms of transfer, the combination of chain-based and term walk vectorization produces the best results. This provides evidence that these features are reasonably domain-general (though, we again note that, while M2k and MPTP are different datasets and may require different reasoning strategies, they are both generated from Mizar). Notably, there was only a 11\% loss of performance when training on MPTP and testing M2K, and a 3\% loss when training on M2k and testing MPTP2078. From this experiment we conclude that cheap structural features, though hand-engineered, are relatively domain-general, which again will make them tough for more complex approaches to beat.

%%%%%%%%%%%%%%%%%%%%%%%%%%%%%%%%%%%%%%%%%%%%%%%%%%%%%%%%%%%%%%
\section{Related Work}

%A coarse overview of learning approaches for ATPs was given in the introduction.
Several approaches have focused on the sub-problem of premise selection (i.e., finding the axioms relevant to proving the considered problem) \cite{AlamaHKTU-jar14-premises-corpus-kernel,Blanchette-jar16-premises-isabelle-hol,Alemi-NIPS16-deepmath,WangTWD-NIPS17-deepgraph}. 
As is often the case with automated theorem proving, most early approaches were based on manual heuristics \cite{hoder2011sine,roederer2009divvy} and traditional machine learning \cite{AlamaHKTU-jar14-premises-corpus-kernel}; though some recent work has used neural models \cite{Alemi-NIPS16-deepmath,WangTWD-NIPS17-deepgraph,rawsondirected2020,crouse2019improving}. 
Additional %a lot of 
research has used learning to support interactive theorem proving \cite{Blanchette-jar16-premises-isabelle-hol,Bancerek-JAR2018-mml}.

Some early research applied (deep) RL %reinforcement learning
for guiding inference \cite{TaylorMSW-FLAIRS07-cyc-rl}, planning, and machine learning techniques for inference in relational domains
\cite{surveyRLRD}. 
Several papers have considered propositional logic or other decidable FOL fragments, which are much less expressive compared to \trail. %compared to ours \cite{KuYaSa-CoRR18-intuit-pl,LeRaSe-CoRR18-heuristics-atp-rl,CheTi-CoRR18-rew-rl}.
Closer to \trail are the approaches described in
\cite{KalUMO-NeurIPS18-atp-rl,zombori2020prolog} where RL is combined with Monte-Carlo tree search
for theorem proving in FOL. However they have some limitations: 
% MAX - I'm taking out this point because they use feature hashing
%1)~%Generalization issues: 
%Their input axioms are represented by features that depend on the vocabulary (i.e., user-defined predicates etc.). As a result, their approaches would not transfer well to new problems with a different vocabulary. 
%Reasoner specific issues: 
1)~Their approaches are specific to tableau-based reasoners and thus not suitable for theories with many equality axioms, which are better handled in the superposition calculus \cite{bachmair1994refutational}, and 
2)~They rely upon simple linear learners and gradient boosting %such as LIBLIN-EAR \cite{chen2016xgboost} and the XGBoost gradient boosting toolkit \cite{chen2016xgboost} (used with the linear regression objective)
as policy and value predictors. 

Our work also aligns well with the recent proposal of an API for deep-RL-based interactive theorem proving in HOL Light, using imitation learning from human proofs \cite{BaLoRSWi-CoRR19-holist}. That paper also describes an ATP as a proof-of-concept. However, their ATP is intended as a baseline and lacks more advanced features like our exploratory learning.

Non-RL-based approaches using deep-learning to guide proof search include \cite{ChvaJaSU-CoRR19-enigma-ng,LoosISK-LPAR17-atp-nn,paliwal2019graph}. These approaches differ from ours in that they seed the training of their networks with proofs from an existing reasoner. In addition, they use neural networks during proof guidance to score and select available clauses with respect \emph{only} to the conjecture. Recent works have focused on addressing these two strategies. For instance, \cite{piotrowski2019guiding} explored incorporating more than just the conjecture when selecting which inference to make with an RNN-based encoding scheme for embedding entire proof branches in a tableau-based reasoner. However, it is unclear how to extend this method to saturation-based theorem provers, where a proof state may include thousands of irrelevant clauses. Additionally, \cite{aygun2020learning} investigated whether synthetic theorems could be used to bootstrap a neural reasoner without relying on existing proofs. Though their evaluation showed promising results, it was limited to a subset of the TPTP \cite{tptp} that excluded equality. It is well known that the equality predicate requires much more elaborate inference systems than resolution \cite{bachmair1998equational}, thus it is uncertain as to whether their approach would be extensible to full equational reasoning.

%\ibrahim{add Urban's recent work including ~\cite{zombori2020prolog}}
%Our work is also inspired by AlphaZero~\cite{alphazero} and prior work on games and tree search~\cite{thinkingfastandslow,alphazero,Silver-nature16-go,Silver-nature17-go} where we randomly explore the search space as AlphaZero (but without MCTS) for the \textit{tabula rasa} training. % regimen. 

\section{Conclusions}
\label{sec:conc}

In this work we introduced \trail, a system using deep reinforcement learning to learn effective proof guidance in a saturation-based theorem prover. \trail outperformed all prior RL-based approaches on two standard benchmark datasets and approached the performance of a state-of-the-art traditional theorem prover on one of the two benchmarks.

{
\small
\bibliographystyle{aaai}
\bibliography{references}
}
% \pagebreak[4]

% --------------
\pagebreak[1]
 \appendix
\section{Appendix}
\label{sec:appendix}

\subsection{Hyperparameter Tuning and Experimental Setup}
\label{sec:hyperparams}
We  used  gradient-boosted tree  search from
scikit-optimize\footnote{\url{https://scikit-optimize.github.io/}} to find effective hyper-parameters using 10\% of the Mizar dataset. This returned the hyperparameter values in Table~\ref{tab:hyperparameters}. The maximum time limit for solving a problem was 100 seconds. Hyper-parameter tuning experiments were conducted over a cluster of 19 CPU (56 x 2.0 GHz cores \& 247 GB RAM) and 10 GPU machines (2 x P100 GPU, 16 x 2.0 GHz CPU cores, \& 120 GB RAM) over 4 to 5 days.% (for hyper-parameter tuning, we added 5 CPU and 2 GPU machines).

Once the best hyperparameters were found, we ran TRAIL and its competitors (see Section 4.1 and Appendix~\ref{subsec:plcop}) on a CPU machine with 56 x 2.0 GHz cores \& 247 GB RAM. At the end of each iteration, collected training examples were shipped to a dedicated GPU server (with 2 x P100 GPUs, 16 x 2.0 GHz CPU cores, \& 120 GB RAM) which trains and updates \trail's policy network.

\begin{table}[h]
\centering
\begin{tabular}{ll}
\toprule
Parameter                  & Value \\
\midrule
Chains Patterns            & 500 \\
Sub-walks                  & 2000 \\
$k$ layers                 & 2     \\
units per layer            & 161   \\
dropout                    & 0.57  \\
$\lambda$ (reg.)           & 0.004 \\
$2d$ (sparse vector size)  & 645   \\
$\tau$ (temp.)              & 1.13 \\
$\tau_0$ (temp. threshold)             & 11  \\
embedding layers           & 4  \\
dense embedding size        & 800  \\
% $\rho$ (expert decay)      & 0.75    \\
reward normalization       &  (i) normalized by difficulty\\

\bottomrule
\end{tabular}
\caption{Hyperparameter values} 
\label{tab:hyperparameters}
\end{table}

\begin{table*}[h!]
\centering
\begin{tabular}{lcccccccccc}
\toprule
         & 1 & 2 & 3 & 4 & 5 & 6 & 7 & 8 & 9 & 10 \\
         \midrule
M2k &   1,042	& 1,266 &	1,428 &	1,490 &	1,507 &	1,527 &	1,533 &	1,549 &	1,552 &	\bf 1,561 \\
MPTP2078      & 363	& 552 &	680	& 730 &	806 &	844	& 858 &	878	& 877 &	\bf 910    \\
\bottomrule
\end{tabular}
\caption{\trail's performance across iterations}
\label{tab:trail_iter}
\end{table*}

\subsection{Underlying Reasoner}
\label{sec:underlying_reasoner}
The current implementation of \trail uses Beagle~\cite{Beagle2015} as its underlying inference execution system. This is purely an implementation choice, made primarily due to Beagle's easily modifiable open source code and friendly license. 
The purpose of Beagle in \trail is only to execute the actions selected by the \trail learning agent; i.e., Beagle’s proof guidance was completely disabled when embedded as a component in TRAIL and whenever Beagle reaches a decision point, it delegates the decision to TRAIL's policy to decide the next action to pick.

Using an off-the-shelf reasoner (like Beagle) as a reasoning shell is to ensure that the set of inference rules available to \trail are both sound and complete, and that all proofs generated can be trusted. % \trail only assumes its underlying reasoner to be saturation-based and is otherwise not reasoner-dependent, i.e., and any reasoner that can apply FOL inference rules can serve the same role as Beagle in \trail. 
Beagle only executes the actions selected by \trail and can thus be replaced by any saturation-based reasoner capable of applying FOL inference rules.

\subsection{\trail's Learning}
Table \ref{tab:trail_iter} shows the performance of {\trail} across iterations. Compared to the first iteration, \trail managed to solved 547 more problems on MPTP2078 and 519 more problems on M2k. This indicates that \trail is learning rather quickly, beating rlCop and plCop on both datasets by the fifth iteration of learning. Interestingly, \trail's performance monotonically increases over the iterations, which indicates that it is not overfitting to a particular subset of the problems within either dataset.

\begin{figure}%[h]
\begin{center}
\includegraphics[width=1.0\columnwidth]{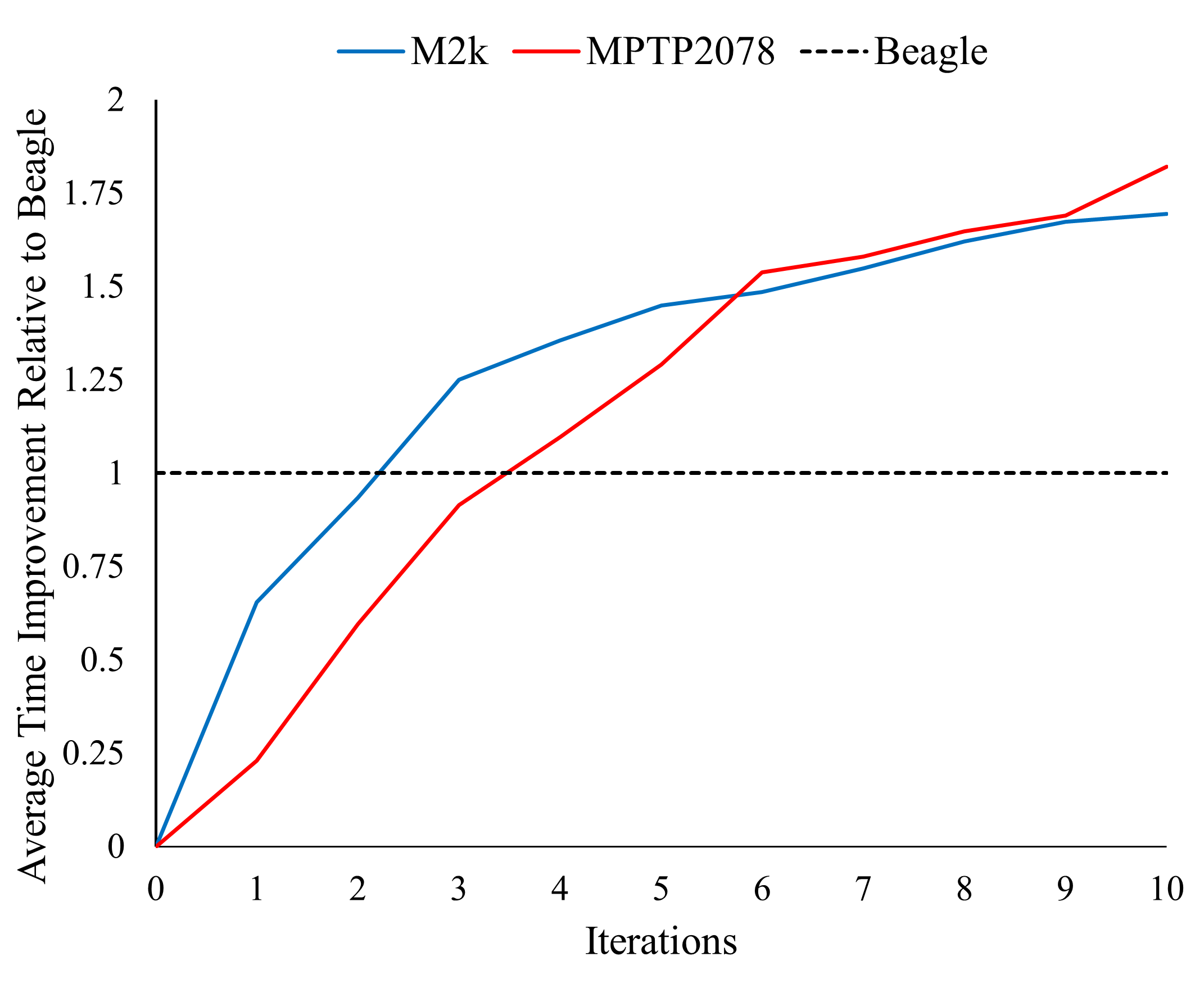}
\vspace{-0.2in}
\caption{\trail's average proof time improvement relative to Beagle (i.e., Beagle's average time to find a proof divided by \trail's average time to find a proof)}
\label{fig:proof_time}
\end{center}
\end{figure}

We also show in Figure \ref{fig:proof_time} the speed at which \trail finds a proof as compared to Beagle. As can be seen in the figure, \trail surpasses Beagle's speed rather quickly (at the second iteration for M2k and the fourth iteration for MPTP). 
One possibility for this is that \trail is initially solving an easier subset of problems, as evidenced by the fact that at both of those iterations, \trail is actually solving fewer problems than Beagle (see Table \ref{tab:trail_iter}). However, by the time \trail reaches iteration 8 on M2k, it solves more problems than Beagle with a 1.6x improvement in terms of time. Similarly on MPTP2078, \trail iteration 5 managed to solved more than 160 problems than Beagle in 1.29x better time. 
% This is interesting, as \trail actually solves fewer problems than Beagle (see Table \ref{tab:trail_iter}) at both of those iterations (in fact, \trail does not surpass Beagle in terms of the number of solved problems on M2k until iteration 8).

\begin{table}
\centering
\footnotesize
\resizebox{1\columnwidth}{!}{
\begin{tabular}{l l l l l l}
\toprule
 & & M2k & Stat. Sig. & MPTP2078 & Stat. Sig. \\
\toprule
\multirow{3}{*} {\setstackgap{S}{4.05ex}\Centerstack[l]{Traditional}}& E & \bf 1922 & \checkmark  (\footnotesize{$z$=-16.9})& \bf 998 & \checkmark (\footnotesize{$z$=-2.7)})\\
% The value of z is -16.9011. The value of p is < .00001. The result is significant at p < .05.
% The value of z is -2.7393. The value of p is .00614. The result is significant at p < .05.
&Beagle & 1543 & & 742 & \checkmark (\footnotesize{$z$=5.3})\\
% The value of z is 0.6803. The value of p is .4965. The result is not significant at p < .05.
% The value of z is 5.3251. The value of p is < .00001. The result is significant at p < .05.
& mlCop & 1034 & \checkmark (\footnotesize{$z$=17.4}) & 502 & \checkmark (\footnotesize{$z$=13.4})\\
% The value of z is 17.4235. The value of p is < .00001. The result is significant at p < .05.
% The value of z is 13.3625. The value of p is < .00001. The result is significant at p < .05.
\midrule
\multirow{3}{*} {\setstackgap{S}{4.05ex}\Centerstack[l]{RL-Based}}& rlCop &  1235 & \checkmark (\footnotesize{$z$=11.2}) & 733 & \checkmark (\footnotesize{$z$=5.6}) \\
% The value of z is 11.2114. The value of p is < .00001. The result is significant at p < .05.
% The value of z is 5.6156. The value of p is < .00001. The result is significant at p < .05.
& plCop & 1359 & \checkmark (\footnotesize{$z$=7.2}) & 782 & \checkmark  (\footnotesize{$z$=4.0})\\   
% The value of z is 7.1748. The value of p is < .00001. The result is significant at p < .05.
% The value of z is 4.0414. The value of p is < .00001. The result is significant at p < .05.
&\trail  & \bf{1561} & & \bf{910} & \\
\bottomrule
\end{tabular}
}
\caption{Number of problems solved in M2k and MPTP2078, best two approaches in \textbf{bold}. Statistically significant differences ($p$ < .05) {\emph{relative to \trail}} are marked with \checkmark.
}
\label{tab:m2k_2078_stat}
\end{table}

\subsection{Statistical Significance Tests}
Table \ref{tab:m2k_2078_stat} shows the performance of \trail compared to learning and traditional theorem provers. This table repeats the results obtained in Table 1 with statistical significance tests ($p < 0.05$) relative to \trail. For example, state-of-the-art traditional theorem prover E outperforms all other approaches including \trail. E outperforms \trail in a statistically significant way with $z = -16.9$ on M2k and $z = -2.7$ on MPTPT2078 dataset. On the other hand, \trail outperforms Beagle in a non-significant way on M2k ($z = 0.6$) and in a significant way on MPTP2078 ($z = 5.3$). Furthermore, all \trail's improvements over mlCop, rlCop and plCop are statistically significant.

\subsection{rlCop and plCop Experiments}
\label{subsec:plcop}

% plcop with paramodulation 		--> M2k: 1301, 2078b: 773
% plcop without paramodulation --> M2k: 1222, 2078b: 707
% rlcop with paramodulation  	 --> M2k: 1238, 2078b: 563
% rlcop without paramodulation 	 --> M2k: 1148, 2078b: 543

\begin{table}[b]
\centering
\begin{tabular}{lcc}
\toprule
      & M2k & MPTP2078 \\
\midrule
TRAIL &   \bf 1,561  &    \bf 910     \\
rlCop (w/o paramodulation) &  1,148   &  543        \\
rlCop (w/ paramodulation) & 1,238 & 563 \\
plCop (w/o paramodulation) & 1,222  & 707 \\
plCop (w/ paramodulation)&   1,301  &    773\\     
\bottomrule
\end{tabular}
\caption{plCop and rlCop performance using same hardware and time limit (100 seconds) as \trail }
\label{tab:rlcop_plcop}
\end{table}

As mentioned in Section 4.1, the numbers reported in Table 1 for plCop and rlCop are taken from their papers \cite{KalUMO-NeurIPS18-atp-rl,zombori2020prolog}. We also replicated their performance under our exact hardware and time constraints. In particular, we used the authors' source code available at \url{https://github.com/zsoltzombori/plcop} which contains the implementation of both rlCop and plCop.  
We used the same default parameters from plCop's configuration files. We noticed, however, that they have two prominent configurations for each dataset (with and without paramodulation) and as a result we decided to report both configurations on each dataset. Table~\ref{tab:rlcop_plcop} shows the performance of \trail, plCop and rlCop on the same hardware with 100 seconds time limit. plCop performance is comparable to what we have in Table~1 while rlCop numbers are lower. 
To avoid any confusion, we decided to use the best performance for both rlCop and plCop in Table 1 which is what the authors reported in their paper. This experiment is to show that the hardware used and the time limits are comparable and hence fair comparison can be made.

% \input{sections/appendix.tex}

% {
% \small
% \bibliographystyle{aaai}
% \bibliography{references}
% }

\end{document}